%% file: main.tex
\documentclass[letterpaper, 10 pt, conference]{ieeeconf}  

\IEEEoverridecommandlockouts                              

\usepackage{graphics} 
\usepackage{epsfig} 
\usepackage{mathptmx} 
\usepackage{times} 
\usepackage{amsmath} 
\usepackage{amssymb}  
\usepackage{xcolor}         
\usepackage{soul}           
\usepackage{multirow}
\usepackage{hyperref}
\usepackage{mathrsfs}
\usepackage{svg}

\title{\LARGE \bf
Can ChatGPT Enable ITS? The Case of Mixed Traffic Control via Reinforcement Learning
}

\author{Michael Villarreal, Bibek Poudel, Weizi Li
\thanks{Michael Villarreal, Bibek Poudel, and Weizi Li are with the Min H. Kao Department of Electrical Engineering and Computer Science, University of Tennessee, Knoxville, Knoxville, TN 37996, USA 
{\tt\small \{tvillarr, bpoudel3\}@vols.utk.edu, weizili@utk.edu}}%
}

\begin{document}

\maketitle
\thispagestyle{empty}
\pagestyle{empty}


\begin{abstract}

The surge in Reinforcement Learning (RL) applications in Intelligent Transportation Systems (ITS) has contributed to its growth as well as highlighted key challenges. However, defining objectives of RL agents in traffic control and management tasks, as well as aligning policies with these goals through an effective formulation of Markov Decision Process (MDP), can be challenging and often require domain experts in both RL and ITS. 
Recent advancements in Large Language Models (LLMs) such as GPT-4 highlight their broad, general knowledge, reasoning capabilities, and commonsense priors across various domains. In this work, we conduct a large-scale user study involving 70 participants to investigate whether novices can leverage ChatGPT to solve complex mixed traffic control problems. 
The participants' task is to develop the state space and reward function for three RL mixed traffic control environments, including ring road, bottleneck, and intersection.  
We find ChatGPT has mixed results. 
For intersection and bottleneck, ChatGPT increases number of successful policies by 150\% and 136\% compared to solely beginner capabilities, with some of them even outperforming experts. 
However, ChatGPT does not provide consistent improvements across all scenarios.

\end{abstract}


\input{sections/intro}

\input{sections/related}
\input{sections/prelim} 
\input{sections/study}

\input{sections/results}
\input{sections/conclusion}
\input{sections/acknowledge}







\bibliographystyle{unsrt}
\bibliography{ref}

\end{document}

%% file: sections/intro.tex
\section{INTRODUCTION}

Large Language Models (LLMs) represent a significant advancement in artificial intelligence. Their usefulness as a general-purpose tool is anticipated to 
have a profound societal impact. 
Experiments with LLMs such as GPT-4~\cite{openai2023gpt4, bubeck2023sparks}, LLaMA~\cite{touvron2023llama}, and PALM2~\cite{google2023palm2} demonstrate their strong reasoning and common sense abilities across domains such as math, science, and coding~\cite{openai2023gpt4, bubeck2023sparks, touvron2023llama, google2023palm2, wardat2023chatgpt, jeblick2022chatgpt}. To achieve this success, LLMs leverage Reinforcement Learning from Human Feedback (RLHF)~\cite{ouyang2022training, bai2022training} to solve the alignment problem, i.e., follow user intent by fine-tuning them on human feedback. While there is a strong focus on improving LLMs through Reinforcement Learning (RL), leveraging LLMs to assist RL problems is in nascent stages. 

The RL framework is inherently challenging because it demands an understanding of Markov Decision Process (MDPs). 
Any RL problem must be formulated to a MDP through designing the state, action, reward, and discount-factor among other components~\cite{sutton2018reinforcement}. This is usually a tedious task that requires numerous experimentation and careful analysis, often by a domain expert. Specifically, in research areas such as Intelligent Transportation Systems (ITS), it is challenging for novices to understand and design effective MDPs. 
Moreover, no standard technique exists for creating general-purpose MDPs that will work across environments and satisfy various goals.  

Over the last decade, RL has been adopted to address complex control problems in ITS such as traffic management and autonomous driving~\cite{haydari2020deep, veres2019deep}. 
As autonomous agents, including vehicles and traffic lights, become more prevalent~\cite{SAEJ3016} in ITS, they introduce new challenges and opportunities. 
One emerging topic is mixed traffic control that uses RL-empowered robot vehicles (RVs) to regulate human-driven vehicles (HVs), thus improving the overall traffic flow
~\cite{wu2021flow,yan2021reinforcement, wang2023learning}. 
This surge in research interest has attracted a broader audience to participate in the topic, resulting in a growing demand for creative decision-making and control strategies enabled by RL. 
However, the initial technical barrier, specifically formulating MDP components to align with a control strategy, poses a challenge. LLMs, with their broad knowledge, commonsense priors, and creative capacity, show promise in reducing these barriers and simplifying the process.

\begin{figure}
    \centering
    \includegraphics[width=0.70\linewidth]{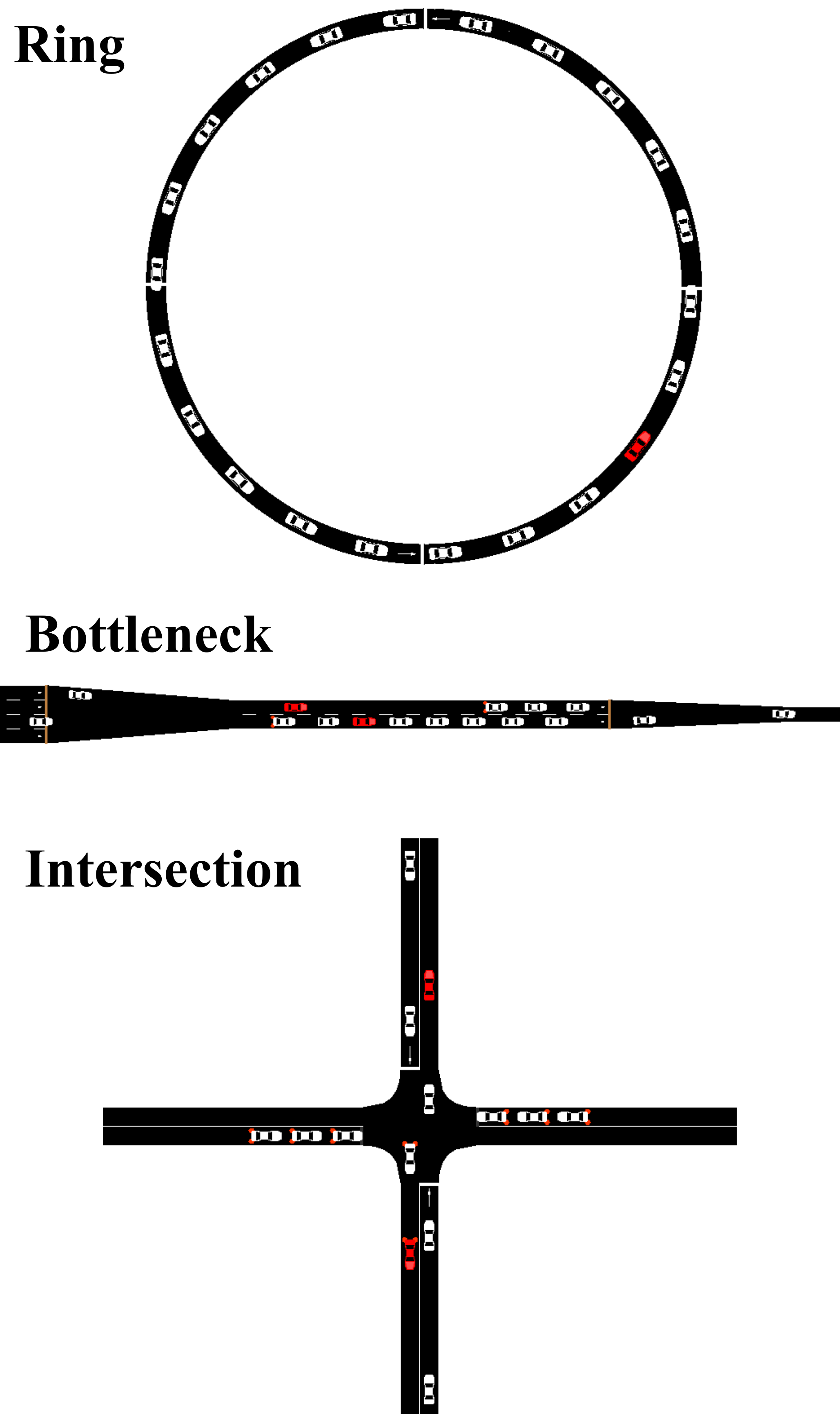}
    \caption{Three mixed traffic control environments~\cite{wu2021flow} (a deep reinforcement learning framework for traffic management), Ring, Bottleneck, and Intersection, are provided to the study participants. Robot vehicles are red and are controlled by learnt RL policies. Human-driven vehicles are white and are modeled by the Intelligent Driver Model~\cite{treiber2000congested}.
    }
    \label{fig:road_envs}
    \vspace{-1.1em}
\end{figure}

In this project, we explore whether ChatGPT (with GPT-4 backend) can assist non-experts in ITS research to solve mixed traffic control problems. 
We conduct a large-scale user study with $70$ participants who have no prior experience in ITS research. The participants are tasked to develop MDP components (state and reward) for three mixed traffic scenarios: a ring road, an intersection, and a bottleneck as shown in Fig.~\ref{fig:road_envs}. We split participants into a control group, where participants attempt to solve problems solely based on common sense and prior knowledge, and a study group, where participants can prompt ChatGPT unrestricted. Participants are provided a manuscript with a general overview of RL, a few examples of MDP components, a reference bank of metrics related to traffic, and images and descriptions of the mixed traffic control environments. The formulated MDPs are then used to train policies and evaluate performance. From the study, we find that:

\begin{enumerate}
    \item{In the intersection environment, using ChatGPT can lead to a performance better than an expert.}
    \item{In the bottleneck and intersection environments, using ChatGPT results in an increase in number of successful policies by $150$\% and $136$\%, respectively.}
    \item{ChatGPT's creativity enables a $363$\% increase in  utilization of new metrics, although the use of these new metrics does not always result in a successful policy.}
    \item{In the ring environment, ChatGPT's help does not increase the policy success rate.} 
\end{enumerate}

The wide range of results observed, from no success to outperforming the expert, across the mixed traffic control environments indicates that the effectiveness of utilizing ChatGPT in ITS depends on the complexity of the specific problem, as well as the extent and manner in which ChatGPT is utilized.

%% file: sections/related.tex
\section{Related Work}

Recent works have studied how Large Language Models (LLMs) can be leveraged in a variety of RL tasks. In robotics, LLMs are utilized in planning and navigation by representing the robotic agent's input (state) in natural language, additionally incorporating the input with visual or raw sensor data for grounding. This approach demonstrates high data efficiency and generalization to unseen environments~\cite{driess2023palm, sashank2023can}. 
Similarly, using LLMs to guide exploration during training can also help achieve greater sample efficiency~\cite{nottingham2023embodied}.

Limited research has been found in minimizing the human effort required to generate an effective policy. Some works use LLMs to substitute components of MDPs, such as making LLM a surrogate or proxy reward function. By using the in-context learning capability of LLMs, hard-to-specify reward functions (e.g., versatility, fairness) have been attempted~\cite{du2023guiding}. Meanwhile, other studies have used LLMs to replace a policy entirely~\cite{yang2023foundation, hu2023language}. However, to the best of our knowledge, no previous work has directly evaluated the effectiveness of LLMs (with and without them) as a tool to reduce the effort in designing MDP components by novices, comparing the obtained policies with those from experts.

%% file: sections/prelim.tex
\section{PRELIMINARIES}
In the following, we introduce the formulation of mixed traffic control as reinforcement learning (RL) tasks and discuss the corresponding test environments.

\subsection{Reinforcement Learning}

We model mixed traffic control as a Partially Observable Markov Decision Process (POMDP) represented by a tuple ($S$, $A$, $P$, $R$, $p_0$, $\gamma$, $T$, $\Omega$, $O$) where $S$ is the state space; $A$ is the action space; $P(s'|s,a)$ is the transition probability function; $R$ is the reward function; $p_0$ is the initial state distribution; $\gamma\in(0, 1]$ is the discount factor; $T$ is the episode length (horizon); $\Omega$ is the observation space; and $O$ is the probability distribution of retrieving an observation $\omega \in \Omega$ from a state $s \in S$. 
At each timestep $t \in [1,T]$, a robot vehicle (RV) uses its policy $\pi_{\theta}(a_t|s_t)$ to take an action $a_t$ $\in$ $A$, given the state $s_t$ $\in$ $S$. The RV's environment provides feedback from taking action $a_t$ by calculating a reward $r_t$ and transitioning the agent into the next state $s_{t+1}$. The RV's goal is to learn a policy $\pi_{\theta}$ that maximizes the discounted sum of rewards, i.e., return, $R_t = \sum^{T}_{i=t}\gamma^{i-t}r_i$. Proximal Policy Optimization~\cite{schulman2017proximal} is used to learn $\pi_{\theta}$.

\subsection{Mixed Traffic Control Environments}
\label{sec:rl_traffic_envs}

\renewcommand{\arraystretch}{1.25}


\subsubsection{Ring} The ring environment (shown in Fig.~\ref{fig:road_envs} top) consists of a single-lane circular road network and $22$ vehicles ($21$ HVs and one RV). It simulates how perturbations due to imperfections in human driving behavior can amplify and propagate, leading to an eventual standstill for some vehicles. This situation, known as `stop-and-go traffic', acts as a wave that propagates continually through the ring, opposite the direction of travel. The RV's goal is to prevent the formation of these waves. Ring is a widely used benchmark in traffic control~\cite{chou2022lord}. An expert's~\cite{wu2021flow} state space (major metrics given in Table~\ref{tab:bank}) is:

\begin{equation}
    \text{s} = \left\{\frac{v_{RV}}{v_{\text{max}}}, \frac{v_{\text{lead}} - v_{RV}}{v_{\text{max}}}, f\left(x_{\text{lead}} - x_{RV}\right)\right\}.
\end{equation}

The difference in $x_{\text{lead}}$ and $~x_{RV}$ is passed through a normalization function $f$. 
An expert's reward function encourages high average velocity and low control actions (acceleration) though a weighted combination given by:
\begin{equation}
    \text{r} = \frac{1}{n}\sum_{i}v_i - \alpha*\left|a_{RV}\right|,
\end{equation}
where $n=22$ and $\alpha$ is chosen empirically.

\subsubsection{Bottleneck} The bottleneck environment (shown in Fig.~\ref{fig:road_envs} middle) simulates vehicles experiencing a capacity drop~\cite{saberi2013empirical} where a road's outflow significantly decreases after the road's inflow surpasses a threshold. The RVs' goal is to improve outflow. 
Bottleneck represents a bridge with lanes decreasing from $4 \times l$ to $2 \times l$ to $l$ (where $l$ is a scaling factor and is one for our work). 
The RV penetration rate is $10\%$. 
An expert's~\cite{vinitsky2018benchmarks} state space is:
\begin{equation}
    \text{s} = \left\{\overline{X}_{HV}, \overline{V}_{HV}, \overline{X}_{RV}, \overline{V}_{RV}, o_{20}\right\}\text{,}
\end{equation}
where the mean positions and velocities of both vehicle types are considered across user-defined segments of the road network. 
An expert's reward function rewards increasing bottleneck outflow: 
\begin{equation}
    \text{r} = o_{10}\text{.}
    \vspace{-0.5em}
\end{equation}

\subsubsection{Intersection}
The intersection environment (shown in Fig.~\ref{fig:road_envs} bottom) represents an unsignalized intersection where east/westbound traffic flow is less than north/southbound traffic flow. This flow discrepancy leads to east/westbound traffic queues as crossing the intersection would be unsafe otherwise. 
RVs drive in the north/south directions with a $20\%$ penetration rate. The RVs' objective is minimizing east/west queues and increased average vehicle velocity. 
An expert's~\cite{vinitsky2018benchmarks} state space is:
\begin{equation}
    \text{s} = \{V_{\text{all}}, I_{\text{all}}, E_{\text{all}}, D_{\text{edge}}, \overline{V}_{\text{edge}}\}\text{,}
\end{equation}
where $I_{\text{all}}$ is all vehicle's distance to intersection, $E_{\text{all}}$ is all vehicle's edge number, $D_{\text{edge}}$ is density of each edge, and $\overline{V}_{\text{edge}}$ is average vehicle velocity of each edge. There are eight edges for each direction on both sides of the intersection.
An expert's reward function penalizes vehicle delay and vehicle standstills in traffic: 
\begin{equation}
    \text{r} = -\frac{t * {\sum}((V_{\text{max}} - V_{\text{all}}) / V_{\text{max}})}{\text{n} + \text{eps}} - (\text{gain} * ss_{n}), 
\end{equation}
where $t$ is current timestep, $n$ is number of vehicles, 
$eps$ prevents zero division, $gain$ is 0.2, and $ss_{n}$ is the number of standstill vehicles.  

\begin{table}[t!]
    \centering
    \begin{tabular}{|p{0.2\linewidth}|p{0.7\linewidth}|}
         \hline
            Type & Metrics \\
         \hline
         Vehicle & 
         Position ($x$), Velocity ($v$), Acceleration ($a$), Gap to leader/follower ($g_{\text{lead}}$/$g_{\text{fol}}$), Fuel consumption ($fc$) \\
         
         \hline
         Road Environment & 
         Outflow over last $i$ seconds ($o_{i}$), Density ($d$), Speed limit ($v_{\text{max}})$, Average speed ($\overline{v}$), Average acceleration ($\overline{a})$, Minimum speed ($v_{\text{min}})$, Average time standstill ($\overline{ss}_{\tau}$) \\
         \hline
         Driving Behavior & 
         Distance/Time to complete stop ($st_{\text{dist}}$/$st_{\tau}$), Time to collide ($ttc$), Jerk ($j$) \\
         \hline
    \end{tabular}
    \caption{The bank of metrics provided to participants. Capitalization of a metric indicates its vector.}
    \label{tab:bank}
    \vspace{-3.0em}
\end{table}

%% file: sections/study.tex
\section{User Study}
\label{sec:methodology}

\subsection{Participants and Cohorts}
\label{sec:user_participants}


\begin{figure*}[h!]
    \centering
    \includegraphics[width=1.0\linewidth]{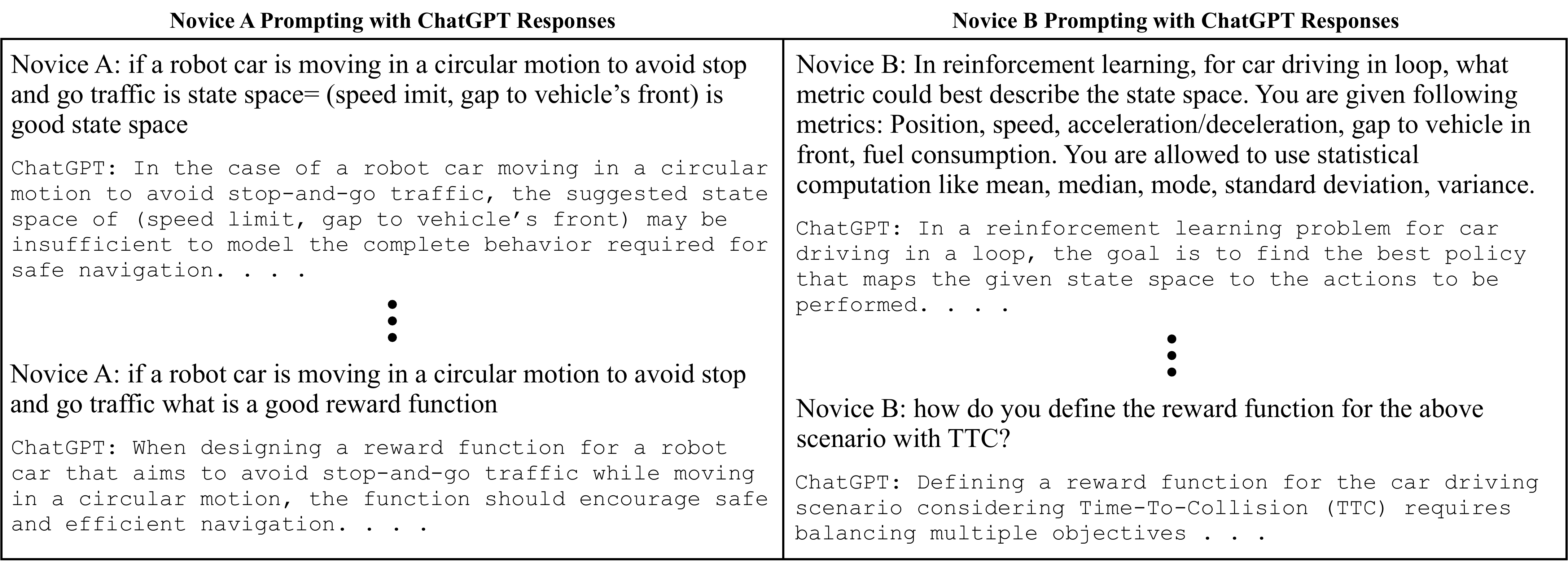}
    \caption{Example prompting by two novices, Novice A and Novice B, with ChatGPT response excerpts for the ring environment. The novices' prompts are given exactly. Only conversation portions are provided for brevity purposes. Novice A, unlike Novice B and most other participants, starts the conversation by asking if a certain state space is good to which ChatGPT responds the state space might be "insufficient." Novice B follows more closely participants' trend to completely rely on ChatGPT for the state space after providing context of the ring environment. Both novices heavily rely on ChatGPT for the reward function; however, Novice B prompts ChatGPT to include time-to-collision based on prior conversation with ChatGPT. 
    }
    \label{fig:prompt_example}
\end{figure*}

\begin{figure*}[h!]
    \centering
    \includegraphics[width=1.0\linewidth]{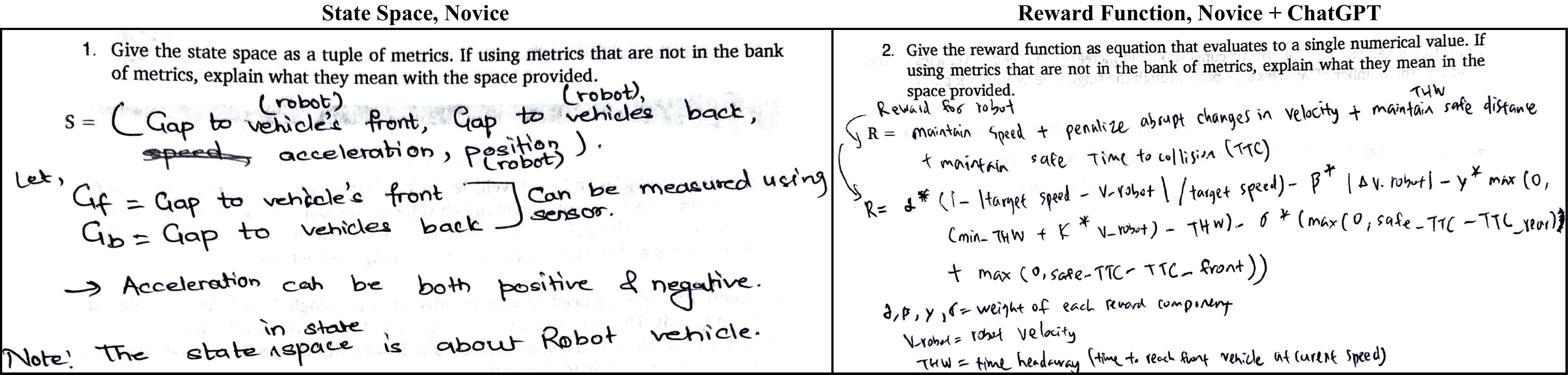}
    \caption{Example state space from novice (left) and reward function from novice with ChatGPT (right) for ring. 
    The novice without ChatGPT's reward function heavily uses provided metrics, while using ChatGPT allows the reward function to contain complex, generated terms. ChatGPT provides sound reasoning for using the reward function metrics (see Sec.~\ref{sec:participant_answers} for details).}
    \label{fig:answer_example}
\end{figure*}

We recruit 70 graduate students as participants ($54$ male, $16$ female). The male/female participants have a median age of $23$/$22$ and average $4.93$/$4.26$ years of driving experience.
Participants come from a machine learning or artificial intelligence course with minimal coverage of reinforcement learning (RL) taught by the same professor. 
At the study's date, artificial intelligence participants have slightly more RL knowledge; however, we find no discernible difference between the two courses' results. We attribute this to the manuscript and ChatGPT's assistance.

\begin{table}[h!]
\centering
\begin{tabular}{|l|c|c|c|c|c|c|}
\hline
 & Control & Study & Control & Study & Control & Study \\
\hline
Valid &  25 & 28 & 24 & 27 & 21 & 24 \\
Invalid &  13 & 4 & 14 & 5 & 17 & 8 \\
\textbf{Total} &  \textbf{38} & \textbf{32} & \textbf{38} & \textbf{32} & \textbf{38} & \textbf{32} \\
\hline
\end{tabular}
\caption{Number of valid  and invalid answers for each traffic environment for the control and study groups.}
\label{tab:valid}
\vspace{-1.0em}
\end{table}




The $70$ participants are split into two cohorts: a control group ($38$ participants) that only use prior non-expert knowledge and the manuscript, and a study group ($32$ participants) that additionally has access to ChatGPT~\cite{openai2023chatgpt, openai2023gpt4}. 
The control group helps assess whether ChatGPT can assist non-experts in solving traffic problems by providing baseline capabilities.
We find the control group heavily uses collision-prevention metrics, including $g_{lead}/g_{fol}\text{,}~ttc\text{,} ~\text{or}~st_{dist}/st_{\tau}$, with consistent use of $x_{RV}\text{,}~v_{RV}\text{,}~\text{and}~a_{RV}$ across all environments. Additionally, the control group tends to include objective-oriented metrics such as $o~\text{or}~\overline{ss}_{\tau}$ for bottleneck and metrics about queue length/time standstill in the east/west directions for intersection.
These metrics are straightforward and intuitive, aligning with our expectations for non-experts in ITS research.



We poll the study group's ChatGPT~\cite{openai2023chatgpt} use frequency
with the following descending frequencies (number of participant selections): daily ($2$), several times a week ($13$), once a week ($6$), several times a month ($5$), once a month or less ($2$), and never ($3$). 


\subsection{ChatGPT Setup}
\label{sec:gpt4_clarification}

The study group participants use GPT-4 with $8$k context length and temperature~$=0.7$, a static model with no additional fine-tuning. The March 14th, 2023 version of GPT-4 is used. To simulate the practical use of ChatGPT, a fresh chat interface is provided to participants using TypingMind~\cite{typingmind}. The participants use the same conversation with ChatGPT for the study to enable in-context learning. 

\subsection{Manuscript}
\label{sec:manuscript}

We develop two versions of manuscripts\footnote{\url{https://github.com/tmvllrrl/its-study}} (for different cohorts) with the following four sections.
Study duration is $95$ minutes. 
Within the $95$ minutes, participants are required to spend the first $25$ minutes reading the first three sections outlined below, and then given the remaining time to look over the entire manuscript. This is to ensure a higher probability of participants reading over the provided material, potentially leading to better quality state spaces and reward functions for the mixed traffic environments.




\subsubsection{General Instructions and Background Questions}
\label{sec:gen_instruct}
The general instructions outline participant conduct for the study's duration. 
and warn participants that rule violations comprise their answers' integrity and result in their dismissal. 
Participants complete the background questions discussed in Sec.~\ref{sec:user_participants}.


\subsubsection{Reinforcement Learning Overview}
The RL overview section provides a brief explanation of the state space and reward function RL components. We provide transportation-related examples and explanations to ensure common understanding of RL across all participants.

\subsubsection{Answer Instructions and Bank of Metrics}
The answer instructions cover how their answers should be formatted with examples.
The bank of metrics is provided in Table~\ref{tab:bank}.
The participants can use any new metric with explanation.  



\subsubsection{Mixed Traffic Environment Descriptions and Questions}
Each of the three traffic environments (shown in Fig.~\ref{fig:road_envs}; details in Sec.~\ref{sec:rl_traffic_envs}) receives a general description, problem explanation, and the mixed traffic objective of the RVs. 
The general description includes details such as the number of RVs present, general flow behavior of traffic, ratio of RVs to human-driven vehicles. 
We provide a supplementary video\footnote{\url{https://youtu.be/qTqgfl76FAo}} 
demonstrating the mixed traffic environments.
We ask the participants three questions per environment: the state space, the reward function, and briefly explain the rationale behind the reward function.
We ask participants to explain their reward function rationale to encourage deeper understanding from the participants.
The state spaces participants provide are observation spaces; however, we solely use ``state space" to prevent additional complexity/avoid confusion.

%% file: sections/results.tex
\section{RESULTS}
Next, we analyze the participants' answers. Then, we briefly discuss experiment setup and the results in the three mixed traffic control environments.

\subsection{Participant Answers Analysis}
\label{sec:participant_answers}


\begin{table}[h!]
\centering
\begin{tabular}{|l|c|c|c|c|c|c|}
\hline
\multirow{2}{*}{Responses} & \multicolumn{2}{c|}{Ring} & \multicolumn{2}{c|}{Bottleneck} & \multicolumn{2}{c|}{Intersection} \\
\cline{2-7}
 & Control & Study & Control & Study & Control & Study \\
\hline
Valid &  25 & 28 & 24 & 27 & 21 & 24 \\
Invalid &  13 & 4 & 14 & 5 & 17 & 8 \\
\textbf{Total} &  \textbf{38} & \textbf{32} & \textbf{38} & \textbf{32} & \textbf{38} & \textbf{32} \\
\hline
\end{tabular}
\caption{Number of valid  and invalid answers for each traffic environment for the control and study groups.}
\label{tab:valid}
\vspace{-1.0em}
\end{table}


We find that ChatGPT~\cite{openai2023chatgpt,openai2023gpt4} impacts provided state spaces/reward functions by significantly decreasing invalid answers over using prior knowledge. 
We consider an answer valid if it contains metrics provided in the metrics bank or have well-defined explanations.
Table~\ref{tab:valid} presents the number of valid/invalid answers for all three environments. On average, only $61$\% of answers from the control group are valid, while $82$\% are valid for the study group, a $21$\% increase. This illustrates ChatGPT's capabilities in guiding participants to create valid state spaces/reward functions.

Another impact of ChatGPT is that the study group uses (in ring, bottleneck, intersection order) $35$, $63$, and $59$ new metrics compared to the control group's 8, $17$, and $21$ new metrics that do not exist in the metrics bank. 
On average, this is a $363$\% increase in new metrics. 
This significant increase implies that ChatGPT can provide new perspectives to solving the mixed traffic problems.

We provide example prompting by two novices (labeled Novice A and Novice B) from the study group for the ring environment's state space and reward function in Fig.~\ref{fig:prompt_example}. Novice A deviates from the general trend of participants by asking ChatGPT if a provided state space is good for the ring environment (to which ChatGPT correctly assumes the state space may be insufficient). The general trend of participants is showcased with Novice B where Novice B provides context for the ring environment, then asks what a state space might be for the environment. Additionally, both novices more closely follow participants' general trend of heavily relying on ChatGPT by asking ChatGPT for a good reward function. Novice B demonstrates a common behavior with their reward function prompting by wanting to have a specific metric (for Novice B, time-to-collision) in the function.

We provide example state space and reward function in Fig.~\ref{fig:answer_example}. The left image is a control group (Novice) state space, while the right image is a reward function from the study group (Novice + ChatGPT). 
The novice state space heavily uses existing metrics (a trend with the control group), while the ChatGPT reward shows the intricate metrics ChatGPT generates. 
ChatGPT also offers explanations that appear reasonable with the terms.
For example, for the ``maintain safe time to collision" in Fig.~\ref{fig:answer_example}, ChatGPT states, ``Since your state space includes the time-to-collision (TTC) for the front and rear vehicles, you can encourage the agent to maintain a safe TTC with both vehicles.... The reward is applied only when the actual TTC is less than the determined $safe TTC$." 

\begin{figure*}[h]
    \centering
    \includegraphics[width=\linewidth]{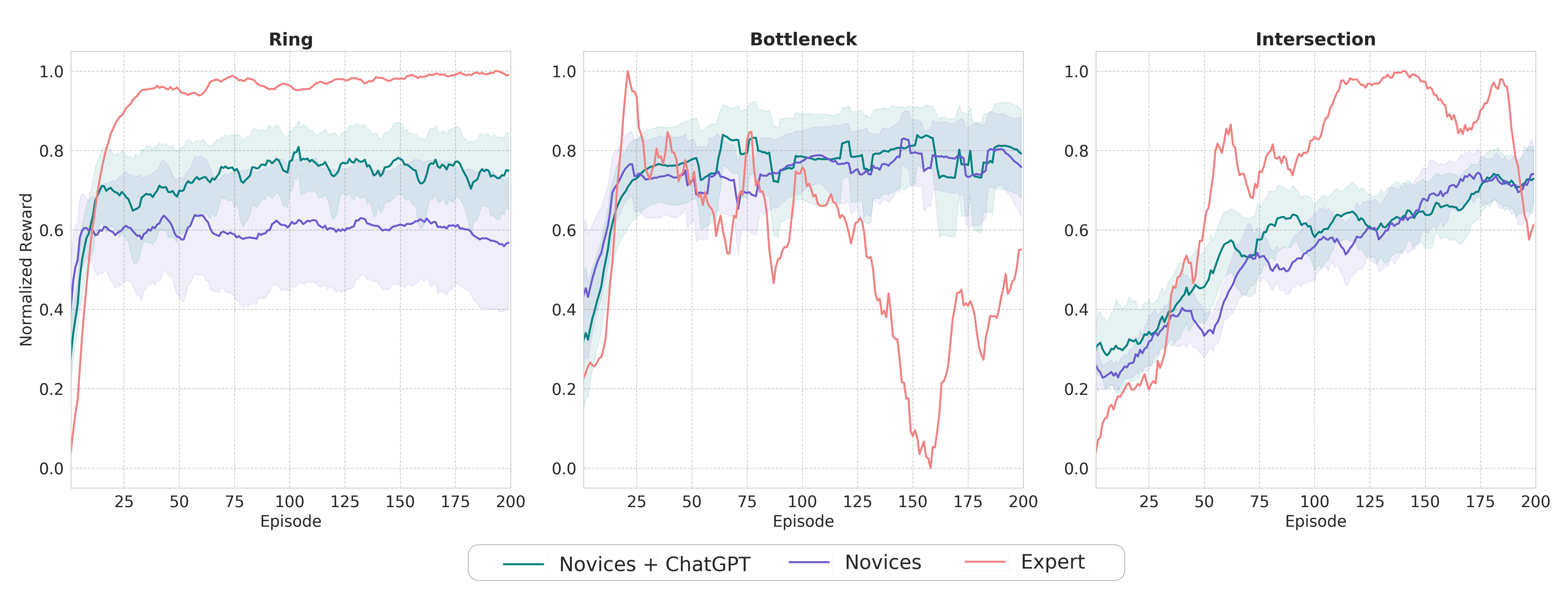}
    \caption{Average normalized reward curves during training are shown for the three traffic control environments. Each environment consists of three curves for the control group (Novices), the study group (Novices + ChatGPT), and an expert. For Novices and Novices + ChatGPT, the solid line indicates the average for all participants, with the shaded region representing variance. In both groups across all networks, average rewards increase over the course of training. This validates that both groups are able to develop state spaces and reward functions that are trainable using RL.
    }
    \label{fig:training_plots}
\end{figure*}

\begin{figure*}
    \centering
    \includegraphics[width=\linewidth]{figures/results_plots.pdf}
    \caption{
    Results for the three mixed traffic environments with successful RL policies denoted. For ring, five policies using ChatGPT's help are successes, while seven policies are successful using only non-expert knowledge.
    Using ChatGPT sees a decrease in successful policies by two compared to only using non-expert knowledge, illustrating ChatGPT needs to be better prompted or further improvements to be useful in this task. 
    For bottleneck, 14 policies are successful without ChatGPT's assistance, while 19 are successful without ChatGPT, a $136$\% increase. 
    For intersection, only using non-expert knowledge results in four successful policies, while ChatGPT increases successes to six (a $150$\% increase) with four (the rightmost green markers) of those outperforming the expert. 
    The bottleneck and intersection increases illustrate how ChatGPT can enable more non-experts to solves complex mixed traffic control tasks. However, the number of increases is lower than expected, potentially showing ChatGPT needs better prompts or further improvement.
    }
    \label{fig:results}
\end{figure*}

\subsection{Experiment Setup}

We train an RL policy for each valid answer 
using Proximal Policy Optimization (PPO)~\cite{schulman2017proximal} with default RLlib hyperparameters~\cite{liang2018rllib}. The HVs use the Intelligent Driver Model (IDM)~\cite{treiber2000congested} with the stochastic noise range [$-0.2$, $0.2$] added to account for heterogeneous driving behaviors. Policies are trained for $200$ episodes.
A fully-connected neural network with 2 hidden layers of size $8$ are used for the ring and intersection environments and size $16$ are used for the bottleneck environment. Experiments are conducted with Intel i7-12700k CPU and $32$G RAM. 

\subsection{Experiment Results}
\label{sec:exp_results}

\subsubsection{Training}

We supply training curves for control (Novices) and study (Novices + ChatGPT) groups in Fig.~\ref{fig:training_plots}. The curves are normalized to [0,1] with the mean reward values averaged across the two respective groups at each episode. We also supply the expert's training curve. 
For all three environments and both groups, the curves show reward improving during training, validating participants were able to develop trainable policies as a result of the RVs' actions. 
Increasing rewards does not guarantee the RV achieves the environment's objective, as the RV may pursue actions that enhance rewards not in line with the goal.

\subsubsection{Ring}

Results are given in Fig.~\ref{fig:results} (left). Due to the task's complexity, we plot a policy's best performance over 10 tests for each trained ring policy. All vehicle average speed (x-axis; meters/second) and minimum speed of any vehicle in the ring (y-axis) for the last 100 seconds (total testing period is 600 seconds) are considered. We consider policies successful (outlined) if their minimum speed is greater than zero while maintaining relatively good average speed. An expert's performance~\cite{wu2021flow} is also provided for comparison.

When using ChatGPT's assistance, five policies are successful, while seven policies are successful using only non-expert knowledge. 
This result defies expectations given ChatGPT's ability to increase valid answers and inject new metrics into the state spaces/reward functions for the ring environment. One explanation is despite the addition of new metrics, the metrics do not improve the robot vehicle's ability to prevent stop-and-go traffic. Another conjecture is while the study group is encouraged to use ChatGPT, the level of usage in participants varies from asking ChatGPT a few questions to completely relying on it. This impacts ChatGPT's ability to help the participants.
Additionally, the high number of non-expert successful policies is unexpected. While none of the given policies reach the expert's level, the anticipated number is close to zero given the tasks' complexity. Non-experts have more capability than originally hypothesized.

\subsubsection{Bottleneck}

Fig.~\ref{fig:results} (middle) shows the bottleneck results. Each trained RL policy is tested $10$ times with the average reported. Outflow (x-axis; vehicles/hour) is considered, and an expert's performance is given as a pink, vertical line. We consider a policy successful if the policy's outflow is greater than $1400$. 

While the outflow range between novices and novices with ChatGPT is similar, 
we observe an increase in successful policies when using ChatGPT's aid. The control group has $14$ successful policies, while the study group has $19$ successful policies, a $136$\% increase. Similar to ring, no participant-given policy outperforms the expert being near $100$ shy.

\subsubsection{Intersection}

Intersection results are given in Fig.~\ref{fig:results} (right). The trained RL policy is evaluated $10$ times with average results reported. All vehicle average speed (x-axis; meters/second) and east/westbound queue length (y-axis; number of waiting vehicles) are considered. We plot an expert's performance~\cite{wu2021flow} and consider the nearest neighbors as successes (outlined).
Four policies are successful with only non-expert knowledge, but six policies are successful (a $150$\% increase) when using ChatGPT aid. Additionally, of the six with-ChatGPT-help policies, four of them outperform the expert policy by a significant margin. This result is significant and showcases how ChatGPT can give non-experts the ability to compete with ITS domain experts.
Two non-expert policies outperform the expert policy, an unexpected outcome though half as much as using ChatGPT.

\begin{figure}
    \centering
    \includegraphics[width=\linewidth]{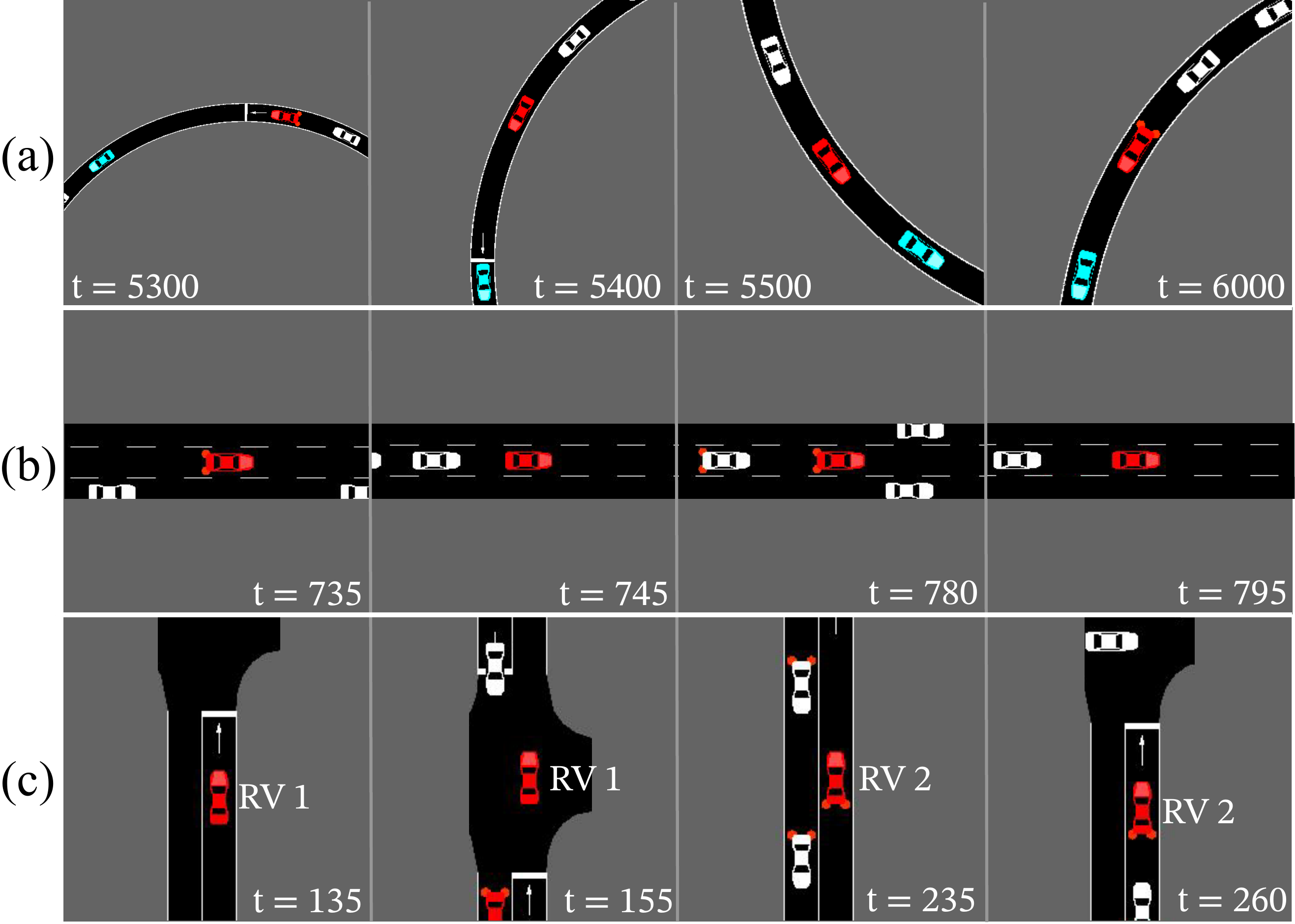}
    \caption{Example successful policies with ChatGPT's aid. (a) \textbf{Ring.} The RV first creates a gap to the leading vehicle (blue) to slow every vehicle down before gradually accelerating to stabilize the ring, preventing stop-and-go traffic. (b) \textbf{Bottleneck.} The RV gradually slows down (before speeding up again in last frame) to force following vehicles to slow down as well, easing congestion closer to the bottleneck's end. (c) \textbf{Intersection.} The first RV travels safely through the intersection, while a second RV slows down so all east/westbound vehicles can cross without causing an incident.}
    \label{fig:frames}
    \vspace{-1.5em}
\end{figure}

For both bottleneck and intersection, we observe a 136\% and 150\% increase, respectively, in successful policies.
Examples of successful policies are provided in Fig.~\ref{fig:frames}. 
ChatGPT is new with its training set not provided meaning ChatGPT could have not been trained on a sufficient amount of RL, ITS, or both training data to provide even more assistance. 
While the level of ChatGPT assistance is participant-determined, we observe a significant number of policies with extensive ChatGPT are not successful. 

 

%% file: sections/conclusion.tex
\section{CONCLUSIONS}
In this work, we conduct a large-scale user study involving non-experts in intelligent transportation systems (ITS) research trying to provide quality reinforcement learning (RL) state spaces and reward functions for three mixed control traffic environments. 
Our study finds using ChatGPT can increase the number of successful polices by $150$\% and $135$\% in the intersection and bottleneck environments, respectively. However, using ChatGPT does not increase successes in the ring environment. Additionally, the improvement rate from using ChatGPT is less than originally theorized. This potentially means that an insufficient amount of RL ITS problems were provided to ChatGPT during training. 


In the future, we intend to expand the study to include more participants, traffic control environments, and Large Language Models (LLMs). A comparative analysis of various LLMs could also be an interesting direction. Further, fine-tuning existing LLMs to be able to serve the ITS community by reducing the human effort required is another avenue that we want to explore.


%% file: sections/acknowledge.tex
\section*{Acknowledgement}
This research is supported by NSF IIS-2153426. The authors would like to thank NVIDIA and the University of Tennessee, Knoxville for their support. The authors would also like to thank Heath Mitchell for GPT-4 API access.